\documentclass[letterpaper, 10 pt, conference]{ieeeconf}  % Comment this line out if you need a4paper

\IEEEoverridecommandlockouts                              % This command is only needed if 
\usepackage{cite}
\usepackage{amsmath, bm}
\usepackage{amssymb,amsfonts}
\usepackage{algorithm}
\usepackage{algpseudocode}
\usepackage{array}
\usepackage{graphicx}
\usepackage{textcomp}
\usepackage{float}
\usepackage{xcolor}
\usepackage{csquotes}
\usepackage{array}
\usepackage{makecell}
\usepackage[pscoord]{eso-pic}
\DeclareMathOperator{\E}{\mathbb{E}}
\DeclareMathOperator*{\argmax}{arg\,max}

\DeclareMathOperator{\KL}{\mathbb{KL}}

\newcolumntype{M}[1]{>{\centering\arraybackslash}m{#1}}

\title{\LARGE \bf
Learning to Forecast Aleatoric and Epistemic Uncertainties \\ over Long Horizon Trajectories}

\author{Aastha Acharya$^{1, 2}$, Rebecca Russell$^{2}$ and Nisar R. Ahmed$^{1}$
\thanks{The authors are with $^{1}$Ann and H.J. Smead Department of Aerospace Engineering Sciences at the University of Colorado Boulder, Boulder, Colorado and $^{2}$The Charles Stark Draper Laboratory, Inc., Cambridge, Massachusetts. {\tt\small aastha.acharya@colorado.edu, rrussell@draper.com, nisar.ahmed@colorado.edu}
}
}

\begin{document}

\maketitle
\thispagestyle{empty}
\pagestyle{empty}

%%%%%%%%%%%%%%%%%%%%%%%%%%%%%%%%%%%%%%%%%%%%%%%%%%%%%%%%%%%%%%%%%%%%%%%%%%%%%%%%
\begin{abstract}
Giving autonomous agents the ability to forecast their own outcomes and uncertainty will allow them to communicate their competencies and be used more safely. We accomplish this by using a learned world model of the agent system to forecast full agent trajectories over long time horizons. Real world systems involve significant sources of both aleatoric and epistemic uncertainty that compound and interact over time in the trajectory forecasts. We develop a deep generative world model that quantifies aleatoric uncertainty while incorporating the effects of epistemic uncertainty during the learning process. We show on two reinforcement learning problems that our uncertainty model produces calibrated outcome uncertainty estimates over the full trajectory horizon.
\end{abstract}

%%%%%%%%%%%%%%%%%%%%%%%%%%%%%%%%%%%%%%%%%%%%%%%%%%%%%%%%%%%%%%%%%%%%%%%%%%%%%%%%
\section{INTRODUCTION}

While there have been major advances in autonomy and reinforcement learning (RL), real world deployment of autonomous agents remains a challenge. Of the numerous factors that contribute to this, arguably the most important one is user trust in the autonomous agent \cite{trust_in_automation, buildingtrust}. Trust depends heavily on user belief in the capabilities of the agent, and establishing appropriate levels of trust can lead to proper usage of and reliance in autonomy. An autonomous agent can help establish trust by self-assessing and communicating its \emph{competency}---its capability and confidence in performing a specified task \cite{Aitken2016AssurancesAM, israelsen}. One way to do this is to have the agent analyze and forecast its own behavior to communicate its competency in the form of a calibrated distribution of outcomes~\cite{acharya2022competency}. 

\begin{figure}
  \centering
  \includegraphics[scale=0.36]{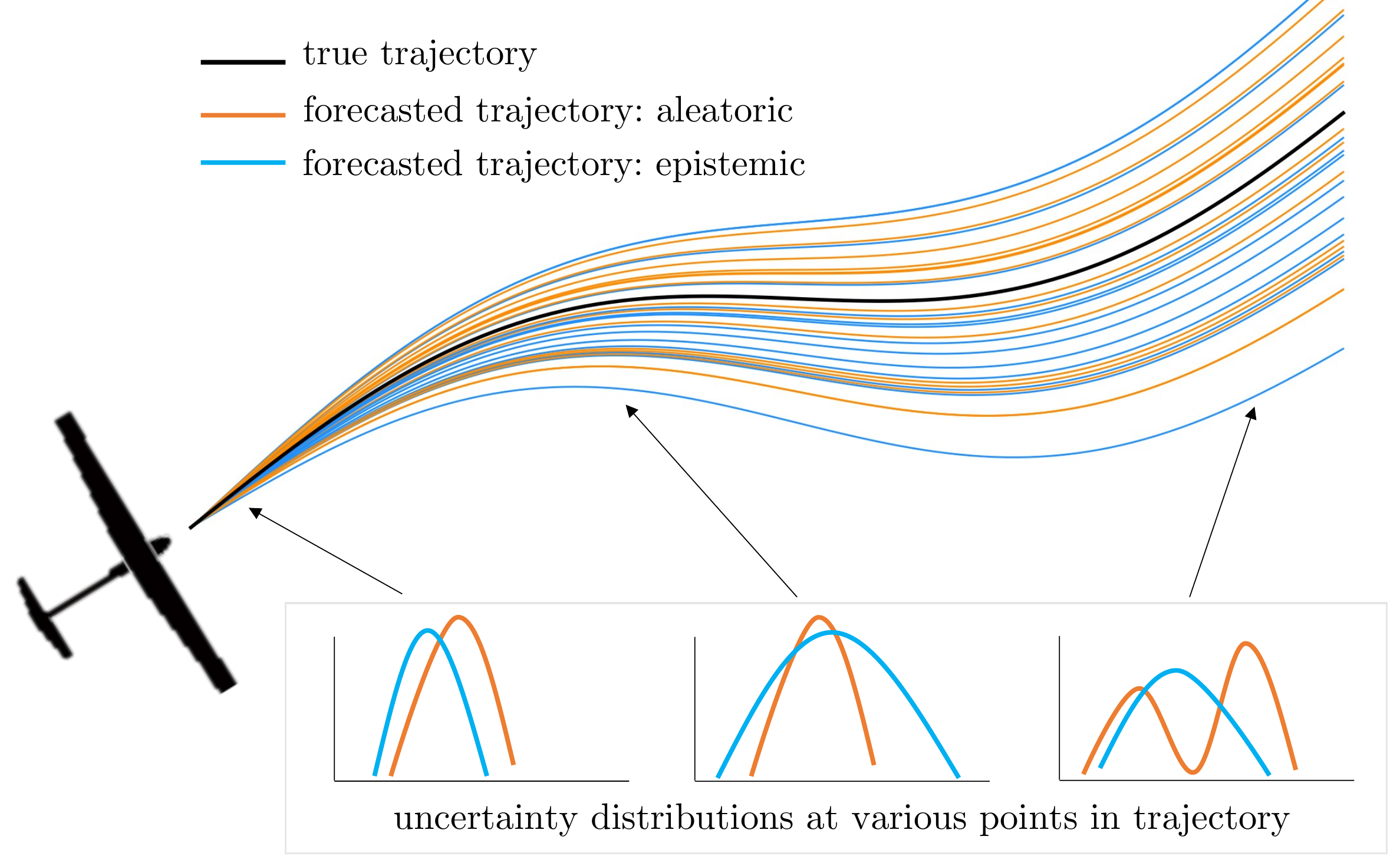}
  \caption{View of forecasted trajectories and its separation into aleatoric and epistemic uncertainty components. Identification and communication of uncertainty components in this way leads to improved competency reporting and outcome assessment by the autonomous agent. }
  \label{overview}
\end{figure}

%In order to 
To communicate actionable information about the outcome distribution, the agent must be %equipped with the ability 
able to perform accurate forecasting over the full time horizon relevant to tasking. Inspired by model-based RL, we use a neural network world model~\cite{ha2018worldmodels} to capture the dynamics and uncertainty associated with the environment, and the agent's interaction with it. Then, we can forecast full trajectories by rolling out the agent's policy with the world model~\cite{acharya2022competency, rosemaryke_longtermfuture}, allowing for competency assessment in multi- or novel task settings. For these models, uncertainty quantification is vital as it ensures that the forecasts are accurate and the generated outcome distributions are trustworthy. Trajectory uncertainties arise from various sources, both from the environment as well as from the modelling process, and can be separated into aleatoric and epistemic components~\cite{chua2018deep, clements_rl_risks_uncertainty}. Our approach, shown in Figure~\ref{overview}, enables the long horizon forecasting over task duration and self-assessment of competency accounting for both components of uncertainty. 

Aleatoric uncertainty~\cite{kendall_gal} is irreducible and arises due to factors such as process noise, measurement noise, and unobserved or partially observed environmental variables. Epistemic uncertainty is reducible with sufficient training data and arises from the modelling process. To quantify epistemic uncertainty, a distribution is placed over the model parameters; to quantify aleatoric uncertainty, a distribution is placed over the model output. Aleatoric uncertainty is naturally intertwined with epistemic uncertainty due to the dependence of model output on the model parameters, making total uncertainty particularly challenging to quantify when both components are significant, as in most real-world environments. For long horizon forecasting, where uncertainty compounds over time, this interplay between aleatoric and epistemic components can dramatically affect the outcome distribution.

In this work, we contribute: (i) a case for combined aleatoric and epistemic uncertainty quantification and its pertinence for real world deployment of competency-aware autonomous agents; (ii) a new method for quantifying aleatoric uncertainty by accounting for any systemic epistemic uncertainty, which is pre-quantified using state-of-the-art ensemble based methods; (iii) results on two different autonomous tasking environments to validate the generality and utility of our approach when there are different levels of epistemic uncertainty in a high aleatoric uncertainty environment. Background information and related works are provided in Sec. \ref{background}. We then describe our technical approach, including a new model training process using deep generative models specifically for aleatoric uncertainty, in Sec. \ref{methods}. Results on two RL robotics environments are provided in Sec. \ref{results}. Finally, we provide conclusions and future work in Sec. \ref{conclusions}.

\section{BACKGROUND AND RELATED WORK}\label{background}

\subsection{Uncertainty Quantification in Deep Learning and RL}

Since the inception of modern deep learning methods, Bayesian deep learning \cite{MacKay92bayesianmethods, MacKay95probablenetworks, neal_bayesian} has received much consideration as a way of quantifying epistemic uncertainty. This comes from the recognition that deep learning models inherently contain uncertainties from their training that must be accounted for, especially when used in applications with real world impact, such as in medical domains \cite{medicalceeplearning2, medicaldeeplearning}, robotic systems \cite{roboticsdeeplearning, roboticsdeeplearning2}, and autonomous vehicles \cite{autdrivingdeeplearning, autdrivingdeeplearning2}. Practical methods such as Monte Carlo dropout \cite{kendall_gal}, deep ensembles \cite{ensembles}, and evidential networks \cite{evidentialnetwork, evidentialnetwork2} approximately quantify epistemic uncertainty. To account for aleatoric uncertainty, entropy-based or variance-based uncertainty separation is typically used \cite{depeweg2017decomposition}. However, these methods do not account for the double counting of epistemic uncertainty that may occur when there is still significant epistemic uncertainty during training, since they do not consider the impact of epistemic uncertainty when modelling the aleatoric uncertainty.

For RL robotics applications that are of particular interest in this work, epistemic uncertainty quantification has been studied extensively~\cite{jain2021deup, osband2021epistemic, eriksson2019epistemic}. This is  justified because many of those applications are in simulation and/or controlled environments which present minimal aleatoric uncertainty. Hence, the focus in these works is instead on 
recognizing insufficient training data or when testing regions are out of bounds from the training range. Out-of-distribution (OOD) or novelty detection is an important class of problems where epistemic uncertainty information is used \cite{sedlmeier2019uncertainty, wu2021uncertainty}. Having OOD knowledge when deployed into the real world can improve the reliability of autonomous agents in novel scenarios. Moreover, having knowledge about the epistemic uncertainty can also aid \enquote{exploration}, especially for RL systems \cite{explorationinRL}. Therefore, modelling of epistemic uncertainty has many benefits for a competency-aware autonomous agent, and this work will rely on epistemic quantification using state-of-the-art ensemble technique \cite{ensembles}. 

Aleatoric uncertainty %and its 
quantification has also been addressed for RL. One primary area of interest is in %recognizing and 
quantifying \enquote{risk} \cite{osband2016risk}, which considers the variance of task returns (e.g. outcome rewards) rather than just the mean \cite{rigter2021risk}. This is useful in recognizing and optimizing for worse case scenarios \cite{neuneier1998risk}. In the context of model-based RL, there are works that have looked at uncertainty propagation by using %uncertainty-aware 
probabilistic dynamics models \cite{chua2018deep}. However, the propagation horizon is fairly short and the authors also assume Gaussian distributions for aleatoric uncertainties. Other works have separately looked at \enquote{perceptual uncertainty}, where known covariance data from sensors are used in trajectory forecasting %process 
\cite{ivanovic2022propagating}. While useful when we have sensor data, this method also makes a strong Gaussian assumption, which will not hold for long horizon forecasting with compounding uncertainties and non-linear behaviors.

In this work, we are interested in creating a competency-aware autonomous agent that can forecast its behavior over long time horizons and communicate the likely outcomes under the provided initial conditions. As such systems expand into the real world, quantifying aleatoric uncertainty becomes just as important as quantifying epistemic uncertainty. This is especially true since competency-aware autonomous agents must forecast and recognize the various conditions that can impact their behaviors and thus their competency over \emph{long time horizons}, where many assumptions---such as Gaussianity---will break down. For example, for an autonomous vehicle that is reporting its competency, the presence of wet driving conditions leads to extra uncertainty in its behavior which manifests itself as aleatoric uncertainty. If the task is to get to a certain location within a designated amount of time, the driving conditions directly contribute to the distribution of outcome traverse times, which determine whether the task can be successfully completed. Thus, one of our primary objectives is to produce \emph{accurate distributions} over the likely outcomes. Since this process of outcome assessment heavily relies on the (correct) distribution of aleatoric uncertainty, we need accurate aleatoric quantification. However, our own experimentation and some past works~\cite{clements_rl_risks_uncertainty} have shown that it is difficult to directly quantify aleatoric uncertainty, as the effects of epistemic uncertainty will naturally be embedded in the process. Therefore, new methodologies are needed to directly account for the epistemic uncertainty in aleatoric model training.

\subsection{Problem Statement} \label{probstatement}

We consider an autonomous agent that interacts with its environment over time, resulting in a trajectory containing state and action information. The states describe both the agent's dynamics along with any relevant environmental conditions, and the actions are those performed by the agent within the environment. Such a trajectory is represented as: 

% \vspace*{-\baselineskip}

\begin{equation}
    \mathcal{\bm{T}} = \{\bm{s}_0, \bm{a}_0, \bm{s}_1, \bm{a}_1, \ldots, \bm{a}_{t-1}, \bm{s}_t, \ldots\},
\end{equation}

\noindent where $\bm{s}_t$ and $\bm{a}_t$ denote states and actions at time $t$, respectively. 
%This trajectory data is useful in categorizing different outcomes and other milestones that may be of interest. %To quantify the uncertainty, The focus here is primarily on quantifying the uncertainties associated with the states. 
Competency assessment requires access to a forecast of all future states up to time $T$ given only the current state and action resulting from any policy. As such, we assume the following distribution is available:

% \vspace*{-\baselineskip}

\begin{equation}\label{eq:traj_prob}
    p(\bm{s}_1, \bm{s}_2, \ldots, \bm{s}_{T}|\bm{s}_0; \bm{a}_0, \bm{a}_1, \dots, \bm{a}_{T-1}),
\end{equation}

\noindent which allows sampling of new states $\hat{\bm{s}}_{t}$ such that, 

% \vspace*{-\baselineskip}

\begin{equation}
    \hat{\bm{s}}_{1}, \ldots, \hat{\bm{s}}_{T} \sim p(\bm{s}_{1}, \ldots, \bm{s}_{T}|\bm{s}_0; \bm{a}_0, \dots, \bm{a}_{T-1}),
    \label{eq:sampletraj}
\end{equation}

\noindent and results in a collection of forecasted trajectories, 

% \vspace*{-\baselineskip}

\begin{equation}
    \mathcal{\bm{T}}_{\textit{forecasted}} = \{\bm{s}_0, \bm{a}_0, \hat{\bm{s}}_{1}, ..., \bm{a}_{T-1}, \hat{\bm{s}}_{T}\}.
\end{equation}

Since we are operating in a model-based RL framework, the learned model of the environment is used to forecast the states out to the desired time horizon $T$. However, there are uncertainties associated with the forecasted states and this is the focus of this work. 

To quantify epistemic uncertainty, the learned model of Equation \ref{eq:traj_prob} is designed such that there is a distribution placed over its parameters. When different parameter values are sampled from this distribution, the model outputs vary accordingly. Hence, given an epistemic uncertainty quantified system with $M$ samples of model parameters, there are $M$ forecasted trajectories $\mathcal{\bm{T}}_{\textit{forecasted}}^{m=1:M}$. The best practice to compute the final prediction is to take an average over all available $m = 1:M$ model outputs. 

For the purpose of aleatoric uncertainty quantification and to formalize our problem, we use a residual to define the difference between the observed trajectories and the corresponding forecasted trajectories. The residual arises due to the uncertainty in the forecasted states and can be separated into the their two components: aleatoric and epistemic. Hence, the total residual $\bm{\epsilon}_{total}$ is defined as: 

\vspace*{-\baselineskip}

\begin{equation} \label{total_res_eps}
    \bm{\epsilon}_{total} = \bm{\epsilon}_{alea} + \bm{\epsilon}_{epist}, 
\end{equation}

\noindent where $\bm{\epsilon}_{alea}$ and $\bm{\epsilon}_{epist}$ denote contributions from aleatoric and epistemic uncertainty, respectively. The statistical definition of residual gives the following equation for total residual: 

\vspace*{-\baselineskip}

\begin{equation} \label{total_res_state}
    \bm{\epsilon}_{total} = \bm{y}_{true} - \frac{1}{M} \sum_{m=1}^M \hat{\bm{y}}_m, 
\end{equation}

\noindent where $\bm{y}_{true} = \{ \bm{s}_1, \bm{s}_2, \ldots, \bm{s}_{T} \}$ are states from the observed trajectory $\mathcal{\bm{T}}$ and $\hat{\bm{y}}_m = \{ \hat{\bm{s}}_{1}^m, \hat{\bm{s}}_{2}^m, \ldots, \hat{\bm{s}}_{T}^m \}$ are states from forecasted trajectory $\mathcal{\bm{T}}_{\textit{forecasted}}^{m}$. The second term in Equation \ref{total_res_state} is the averaged value of all the available model output states. We use a temporally compressed version of $\bm{y}_{true}$ and $\hat{\bm{y}}_m$ to simplify the notation. 

To define the epistemic residual, each of the individual model forecasts are compared against the mean of the forecast. This fits the formulation of state-of-the-art uncertainty decomposition methods which present the variance of the individual model outputs as epistemic uncertainty \cite{depeweg2017decomposition}. Hence, the epistemic residual is defined as: 

% \vspace*{-\baselineskip}

\begin{equation} \label{epist_res}
    \bm{\epsilon}_{epist} = \hat{\bm{y}}_m - \frac{1}{M} \sum_{m=1}^M \hat{\bm{y}}_m.
\end{equation}

Then, following (\ref{total_res_eps}), the aleatoric residual $\bm{\epsilon}_{alea}$ is: 

\vspace*{-\baselineskip}

\begin{equation}
\begin{split}
    \bm{\epsilon}_{alea} &= \bm{\epsilon}_{total} - \bm{\epsilon}_{epist} = \bm{y}_{true} - \hat{\bm{y}}_m, 
\end{split}
\end{equation}

\noindent where $\hat{\bm{y}}_m$ is sampled from any of the available $M$ ensemble models. Under this formulation, the aleatoric residual is dependent on both the observed and the forecasted trajectories. 

Our process is to first use existing methods of epistemic uncertainty quantification to place a distribution over model parameters. Then, different model outputs, representing forecasted trajectories, from the sampled model parameters are used in conjunction with the observed trajectory to define aleatoric residual $\bm{\epsilon}_{alea}$ and collect the training data. Using this data, the goal is to learn the distribution: 

\vspace*{-\baselineskip}

\begin{equation}
    p(\bm{\epsilon}_{alea, 1}, \bm{\epsilon}_{alea, 2}, \ldots, \bm{\epsilon}_{alea, T} | \bm{s}_0; \bm{a}_0, \bm{a}_1, \dots, \bm{a}_{T-1}), 
\end{equation}

\noindent where $\bm{\epsilon}_{alea, t}$ represents the aleatoric residual at each time $t$ in a trajectory. We make no assumption regarding the distribution of the aleatoric residuals, and assume the availability of the first state, a policy, and a collection of forecasted trajectories which is used to define $\bm{\epsilon}_{alea, t}$. 

\section{METHODS}\label{methods}

\begin{figure*}
  \centering
  \includegraphics[scale=0.40]{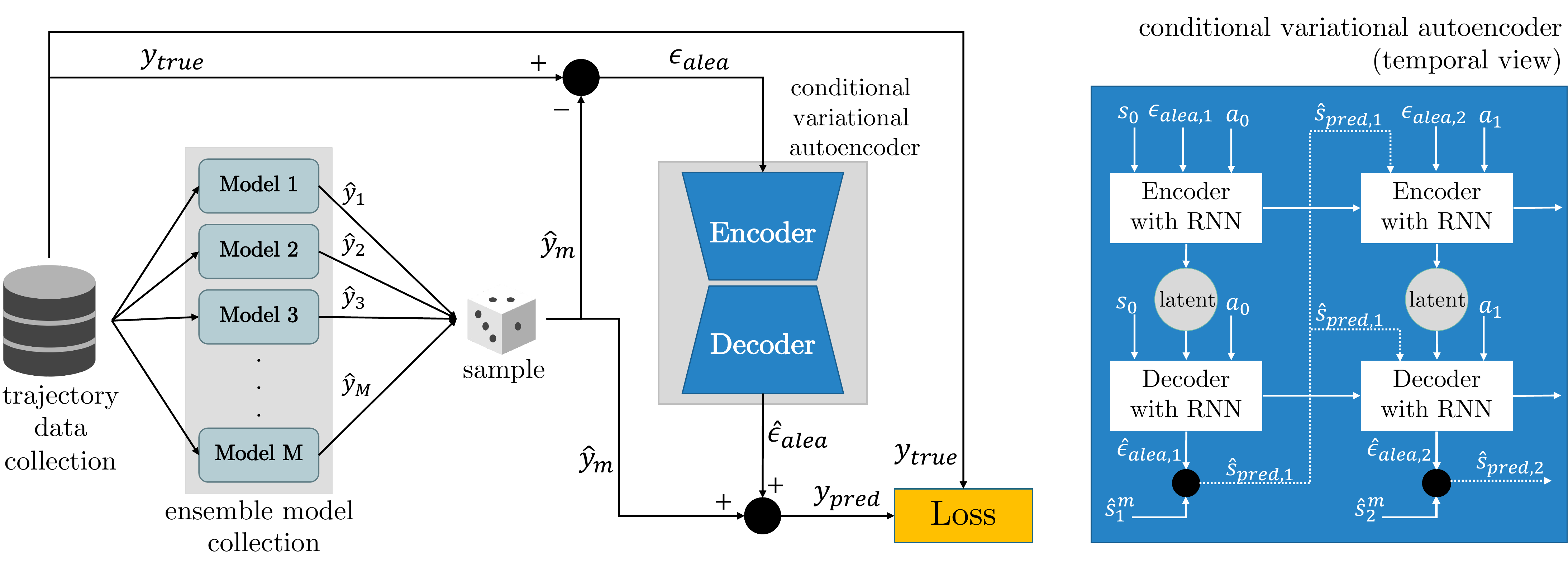}
  \caption{(Left) End-to-end model training process with temporal information excluded for clarity. First, the trajectory data collection is used to train an ensemble of $M$ models. The outputs from $M$ ensemble models are uniformly sampled to select $\hat{\bm{y}}_m$, which is used in conjunction with the training data $\bm{y}_{true}$ to get a residual aleatoric value $\bm{\epsilon}_{alea}$. The conditional variational autoencoder (CVAE) is trained on $\bm{\epsilon}_{alea}$ while the final loss is computed using the prediction $\bm{y}_{pred} = \hat{\bm{\epsilon}}_{alea} + \hat{\bm{y}}_m$ and true observation $\bm{y}_{true}$. (Right) A temporal view of CVAE, which contains a recurrent neural network (RNN) within both the encoder and the decoder to capture temporal memory. This detailed view also shows all the input and output variables, such as the true state $s_t$, actions $a_t$, ensemble model prediction state $\hat{s}_t^m$, and final prediction state $\hat{s}_{pred, t}$.}
  \label{architecture}
\end{figure*}

\subsection{Deep Ensembles for Epistemic Uncertainty}\label{eps_uncertainty}

To quantify epistemic uncertainty, we follow the state-of-the-art method using deep ensembles \cite{ensembles}. The observed trajectory data is used to train $M$ ensemble models, each one designed to forecast states as as defined in (\ref{eq:sampletraj}). Due to the temporal nature of our applications, we train a collection of recurrent neural networks (RNNs), where each one learns the optimal model parameters $\theta^*$ such that:
\begin{equation}
    \theta^* = \argmax_\theta \prod^n_{i=1} p_\theta(\bm{s}^i_{1:T}|\bm{s}^i_0; \bm{a}^i_{0:T-1}),
\end{equation}
\noindent where $n$ is the number of training samples and $\theta$ corresponds to model parameters for a single model that is being trained. By training $M$ ensemble models, $M$ different model parameters $\theta_{m=1:M}$ are obtained. Then, model output corresponding to each of model parameters $\theta_m$ produce a collection of state forecasts $\hat{\bm{y}}_m = \{ \hat{\bm{s}}_{1}^m, \hat{\bm{s}}_{2}^m, \ldots, \hat{\bm{s}}_{T}^m \}$.

\subsection{Deep Generative Models for Aleatoric Uncertainty}

Generative models learn the joint probability distribution of the underlying training data. There are many variations of generative models, and they provide more flexibility than discriminative models in capturing different distribution types. This makes them suitable to model aleatoric uncertainty, especially over long horizon forecasting where complex compounding of uncertainty occurs~\cite{acharya2022uncertainty, huang2022evaluating}. To our knowledge, our work is the first to analyze the usage of generative models for aleatoric uncertainty isolation for long horizon forecasting. 

The desired distribution to learn is $p(\bm{\epsilon}_{alea, 1:T}| \bm{s}_0; \bm{a}_{0:T-1})$, which represents the aleatoric residual as defined in Section \ref{probstatement}. We use a conditional variational autoencoder (CVAE) that uses a latent space to capture the stochasticity of the data \cite{Kingma2014, condVAE}. Since we are working with temporal data, a recurrent structure can be added within both the encoder and decoder portions of the CVAE so that there is memory within the network. An overview of the end-to-end training process alongside a detailed view of the temporal inputs and outputs of the CVAE is shown in Figure \ref{architecture}. 

We use both the observed trajectory data and the forecasted trajectory data from the ensembles models to train the model. During the training process, we simultaneously sample from both of these trajectory sources. Then, aleatoric residual $\bm{\epsilon}_{alea}$ is calculated as the difference between the observed trajectory states and the forecasted trajectory states. By incorporating sampled forecasted states in this way, we directly incorporate the residual contributions from epistemic uncertainty. Note that the aleatoric residual depends on each of the individual model outputs from an ensemble rather than the mean of all the models. This uses the variance information resulting from the variation in model parameters $\theta_m$, providing an indication of epistemic uncertainty \cite{depeweg2017decomposition}.

Although the input and the output from the CVAE is $\bm{\epsilon}_{alea}$ and its prediction $\hat{\bm{\epsilon}}_{alea}$, respectively, the loss function optimizes the difference between the state prediction $\bm{y}_{pred} = \hat{\bm{\epsilon}}_{alea} + \hat{\bm{y}}_m $ and the observed state $\bm{y}_{true}$. This prevents any individual, computed $\bm{\epsilon}_{alea}$ to be considered the \enquote{true} aleatoric residual value. The loss function follows the format of the original CVAE loss \cite{condVAE}, and is described as: 

% \vspace*{-\baselineskip}

\begin{equation}\label{loss}
    \mathcal{L}\left(\bm{\hat{s}}_{pred, 1:T}|\bm{s}_0, \bm{a}_{0:T-1}\right) = \sum_{t = 0}^{T-1} l_t,
\end{equation}
\noindent where

\vspace*{-\baselineskip}

\begin{equation}
\begin{split}
    &l_t = \\ 
    & - \E_{q_{\phi}\left(\bm{z}_{t}|\bm{s}_{0:t}; \bm{a}_{0:t}\right)} \left[\log p_{\theta}\left(\bm{\bm{\epsilon}}_{alea, t+1}|\bm{s}_{0:t} ; \bm{a}_{0:t}; \bm{z}_{t} \right) + \hat{\bm{s}}_{t+1}^m \right] \\
    & \quad + \beta \times \KL\left[ q_{\phi}\left(\bm{z}_{t}|\bm{s}_{0:t} ; \bm{a}_{0:t}\right) || p_{\theta}\left(\bm{z}_{t}\right)\right], 
\end{split}
\label{vae_loss}
\end{equation}

% \vspace*{-\baselineskip}

\noindent and $\phi$ represents the encoder parameters, $\theta$ represents the decoder parameters, $\bm{z}_{t}$ is the latent variable at time $t$, and $q_{\phi}$ is an approximate distribution over those latent variables. The $\beta$ variable in Equation \ref{vae_loss} weighs the KL divergence term, and helps regulate how the two terms in the loss interplay with each other. Inclusion of $\beta$ helps to ensure there is no mode collapse occurring in the CVAE training process. The full training process is further described in Algorithm \ref{alg:aleatoric}.

\begin{algorithm}
{\small{
	\caption{Aleatoric uncertainty residual model training process } 
	\label{alg:aleatoric}
	\begin{algorithmic}[1]
            \State Collect $M$ ensemble models trained on observed trajectory data, resulting in model parameters $\theta_{m=1, \ldots, M}$
            \For {each training batch}
                \State Sample true trajectory states $\bm{y}_{true} = \{ \bm{s}_{1:T} \}$ and \par
                    \hskip \algorithmicindent actions $\bm{A}_{true}=\{ \bm{a}_{0:T-1} \}$
                \State Uniformly sample $m \sim U[0, M]$
                \State Collect forecasted trajectory states $\hat{\bm{y}}_m = \{ \hat{\bm{s}}_{1:T}^m \}$ \par
                    \hskip \algorithmicindent  using $p_{\theta_m}(\bm{s}_{1:T}|\bm{s}_0; \bm{a}_{0:T-1})$
                \State Compute aleatoric residual $\bm{\epsilon}_{alea} = \bm{y}_{true} - \hat{\bm{y}}_m$ 
                \State Train CVAE using $\bm{\epsilon}_{alea}$, $\bm{y}_{true}$, and $\bm{A}_{true}$
                \State Compute loss using CVAE output aleatoric residual  \par
                    \hskip \algorithmicindent $\hat{\bm{\epsilon}}_{alea}$ such that $\bm{y}_{pred} = \hat{\bm{\epsilon}}_{alea} + \hat{\bm{y}}_m $ is compared \par
                    \hskip \algorithmicindent to $\bm{y}_{true}$ 
            \EndFor
	\end{algorithmic} 
 	}}
\end{algorithm}

% \vspace*{-\baselineskip}

\begin{figure*}
  \centering
  \includegraphics[scale=0.39]{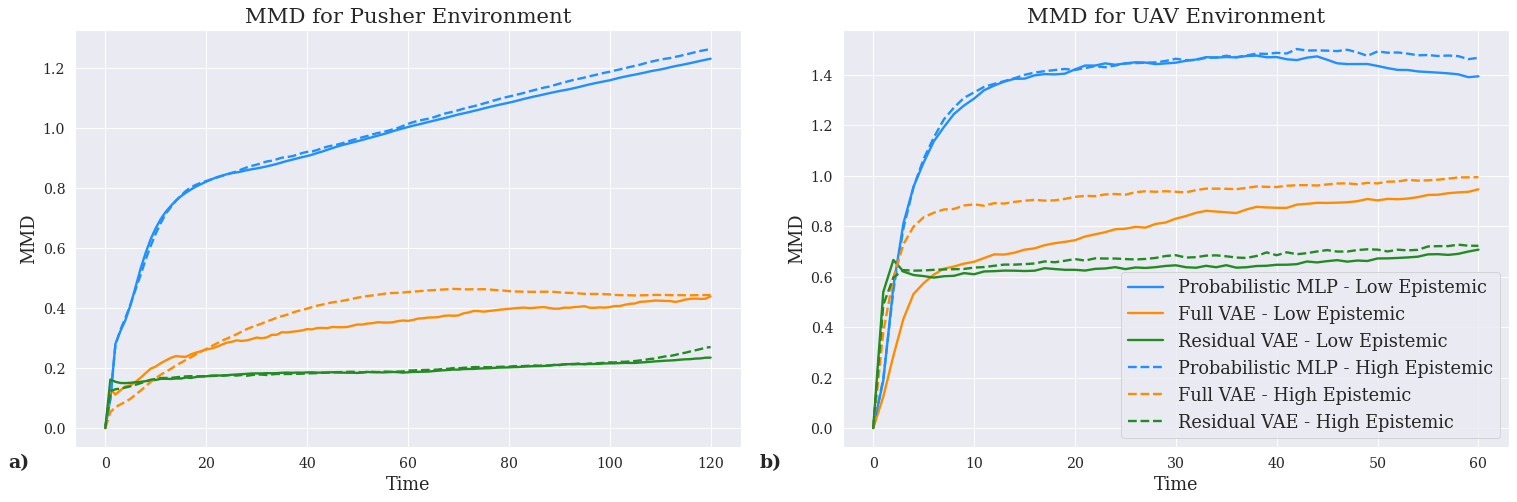}
  \caption{Maximum mean discrepancy (MMD) plots for the Pusher and UAV environment for the three comparison models in low epistemic + high aleatoric (solid line) and high epistemic + high aleatoric (dashed line) scenarios (lower MMD is better). The plots show the effectiveness of residual VAE in achieving the lowest MMD values.}
  \label{mmd}
\end{figure*}

% \vspace*{-\baselineskip}

\subsection{Comparison Models}

To compare the developed residual CVAE model, we train two additional models that quantify aleatoric uncertainty: 

\begin{itemize}
    \item \textbf{Probabilistic MLP} \cite{kendall_gal}: A probabilistic multi-layer perceptron (MLP) which predicts a Gaussian distribution by outputting the mean and the variance. When deployed in a temporal environment, the next state is sampled from the output Gaussian distribution, and used as an input in the following time step. In state-of-the-art uncertainty-aware model-based RL \cite{chua2018deep}, this model is used to capture the aleatoric uncertainty. Hence, comparison against this model demonstrates how our developed method holds against a tried/tested method. 
    \item \textbf{Full VAE} \cite{acharya2022competency}: A recurrent conditional VAE model that is trained on the full training data. This model makes no assumption on the distribution shape of the training data and has a recurrent structure to capture temporal correlations. It is designed for long horizon forecasting of the states but is not designed to isolate aleatoric uncertainty in its quantification process. Comparison against this model shows what a naive approach to aleatoric uncertainty quantification using generative models, without accounting for epistemic uncertainty, looks like. 
\end{itemize}

\subsection{Metrics}

We use the following metrics to compare the models:

\begin{itemize}
    \item \textbf{Maximum mean discrepancy (MMD) with Gaussian kernel}: nonparametric statistical distance metric which compares the forecasted trajectories against the observed. Each forecasted state dimension is compared with the corresponding environmental observation at a temporal level, and the average across all of the states are reported. This is a good indication of aleatoric uncertainty because the observed trajectories' stochasticities come entirely from aleatoric uncertainty sources. Hence, ensuring the distribution of the forecasted trajectories matches the observed trajectories ensures that the aleatoric distribution is being captured accurately.
    \item \textbf{Brier score}: a proper scoring rule that measures the accuracy of predicted outcome probabilities using the mean squared error between the predicted and the actual outcome. This metric directly analyzes the outcome probabilities, which are important for competency-aware autonomous agents. Given that outcome distribution results from the model forecasted states distribution, this is an indirect metric on aleatoric uncertainty distribution that analyzes its impact on downstream outcome analysis. 
\end{itemize}

\section{RESULTS}\label{results}

We present uncertainty forecasting results for two levels of uncertainty (high aleatoric + high epistemic, and high aleatoric + low epistemic) for two different RL environments: Pusher and unmanned aerial vehicle (UAV). 
% The metrics used to compare the results are: 1) maximum mean discrepancy (MMD) with Gaussian kernel and 2) Brier score. MMD is a nonparametric statistical distance metric which compares the forecasted trajectories against the observed. Each forecasted state dimension is compared with the corresponding environmental observation at a temporal level, and the average across all of the states are reported. This is a good indication of aleatoric uncertainty because the observed trajectories' stochasticities come entirely from aleatoric uncertainty sources. Hence, ensuring the distribution of the forecasted trajectories matches the observed trajectories ensures that the aleatoric distribution is being captured accurately. The Brier score is a proper scoring rule that measures the accuracy of predicted outcome probabilities using the mean squared error between the predicted and the actual outcome. This metric directly analyzes the outcome probabilities, which are important for competency-aware autonomous agents. Given that outcome distribution results from the model forecasted states distribution, this is an indirect metric on aleatoric uncertainty distribution that analyzes its impact on downstream outcome analysis. 

\subsection{Pusher}

The Pusher environment is a modified version of the Gym library Reacher environment \cite{brockman2016openai}. To the Reacher environment, we add a ball which has to be pushed to different target locations using the arm (the double pendulum). The state-space for this environment is 12-dimensional: arm position in $(x, y)$, cosine and sine of the upper arm angle $\theta$, the angular rate of the upper arm angle $\dot{\theta}$, cosine and sine of the lower arm angle $\gamma$, angular rate of lower arm angle $\dot{\gamma}$, and ball position and velocity in $(x, y, \dot{x}, \dot{y})$. The 2-dimensional applied action is the torque on the upper and lower arm joints. To create a high aleatoric variation of this environment, we make the Pusher stochastic by injecting Gaussian noise internally using the actions, which over time compounds to create complex, non-Gaussian distributions. To create a high epistemic scenario, we reduce the amount of data available for training by $10 \times$ in comparison to the low epistemic scenario. 

As a first point of comparison, we consider the MMD plots shown in Figure \ref{mmd}a (lower is better). The results are shown for forecasts out to $120$ time steps, and demonstrate that MMD is highest for the probabilistic MLP model. This is expected because the probabilistic MLP is not designed for long horizon forecasting and makes a very strong assumption that the aleatoric distribution is Gaussian for all time. This assumption does not hold for the Pusher environment where there are two entities interacting with each other and causing momentum transfer, bringing forth interesting bifurcating behaviors.
% This is not true because as we go further into the forecasting horizon, there is a higher chance for bifurcation effect showing non-Gaussian behavior in the Pusher environment. Additionally, there are two entities (the arm and the ball) interacting with each other and the momentum transfer between the two can bring forth interesting behaviors. 
The two VAE models, which do not make any assumption about the distribution of aleatoric uncertainty, produce significantly lower MMD values. The results from the full VAE, which does not actively subtract epistemic uncertainty, is higher than with the residual VAE. When analyzing the full VAE based on the epistemic uncertainty level, we see that the high epistemic case typically results in higher MMD. In the case of the residual VAE, both levels of uncertainty show comparable MMD values. Overall, these results show the clear benefit of using the residual VAE for accurate aleatoric quantification for long horizon trajectories.

% Furthermore, we see that the high epistemic case results in the higher MMD for the full VAE. In the case of the residual VAE, it gives the lowest MMD values, with both low epistemic and high epistemic scenarios showing comparable results. These results show the clear benefit of using the residual VAE for accurate aleatoric quantification for long horizon trajectories. 

% \vspace*{-\baselineskip}

\begin{table}[hbt!]
\caption{Brier scores for Pusher environment (lower is better)}
\begin{center}
\renewcommand{\arraystretch}{2}
\begin{tabular}{  | M{1.5cm} || M{1.5cm} | M{1.5cm} | M{1.5cm} | }
 \hline
    & \makecell{Residual \\ VAE} & \makecell{Full \\ VAE} & \makecell{Probabilistic \\ MLP}\\
    \hline \hline
    \makecell{Low \\ Epistemic} & $\bm{0.072}$  & $0.391$ & $0.416$ \\
    \hline \hline
    \makecell{High \\ Epistemic} & $\bm{0.061}$  & $0.439$ & $0.539$ \\
\hline
\end{tabular}
\end{center}
\label{pusher_brier_scores}
\end{table}

% \vspace*{-\baselineskip}

Next, we compute the Brier score relative to the outcomes that are of interest. Although multiple outcomes can be analyzed under our framework, we present results for when the arm is tasked to get the ball to a target location within a given amount of time. The target location, the starting location of the ball, and the starting location of the arm all vary for the 250 analyzed scenarios. The Brier scores for output outcome probabilities are shown in Table \ref{pusher_brier_scores}. The residual VAE produces the lowest value of the Brier scores, indicating that the predictions from this model are the most accurate. Second, the Brier scores produced by the full VAE are also lower than those from the probabilistic MLP, which give the least accurate outcome probabilities.

\subsection{UAV}

The second experimental environment is a Gazebo-based UAV simulation \cite{conlon2022generalizing}. 
% , consisting of the UAV platform model, wind model, temperature model, and battery model. 
The $17$-dimensional state-space for this environment is made up of the UAV platform position $(x, y, z)$,  velocity $(\dot{x}, \dot{y}, \dot{z})$, orientation $(\phi, \theta, \psi)$, and angular velocity in $(\dot{\phi}, \dot{\theta}, \dot{\psi})$. Additionally, the 3-dimensional wind component in $(x, y, z)$ is also included, as is the temperature and the battery percentage. To introduce aleatoric uncertainty into this environment, we include a hidden variable that directly impacts the dynamics of the vehicle but is not part of the state-space representation. This hidden variable is a payload mass, which varies for any given trajectory. Similar to the Pusher environment, we create a high epistemic scenario by limiting the amount of data available for training. 

The UAV MMD results are shown in Figure \ref{mmd}b for $60$ time steps. These results also show probabilistic MLP with the highest and residual VAE with the lowest MMD values. The actual values are larger than the Pusher, which is expected since this is a more complex environment. Additionally, we again see that both the probabilistic MLP and the residual VAE produce similar values for both high and low epistemic scenarios. However, this difference is more pronounced for the full VAE, where the high epistemic scenario results in a higher MMD than the low epistemic scenario. This shows that the full VAE may be double counting epistemic uncertainty as aleatoric uncertainty in this environment. 
% This is the expected behavior because in the high epistemic uncertainty, we would expect the VAE to count epistemic uncertainty as part of its aleatoric uncertainty quantification, thus leading to a greater mismatch between the environmental distribution and the VAE outputted distribution. 

\begin{table}[hbt!]
\caption{Brier scores for UAV environment (lower is better)}
\begin{center}
\renewcommand{\arraystretch}{2}
\begin{tabular}{  | M{1.5cm} || M{1.5cm} | M{1.5cm} | M{1.5cm} | }
 \hline
    & \makecell{Residual \\ VAE} & \makecell{Full \\ VAE} & \makecell{Probabilistic \\ MLP}\\
    \hline \hline
    \makecell{Low \\ Epistemic} & $\bm{0.190}$  & $0.272$ & $0.712$ \\
    \hline \hline
    \makecell{High \\ Epistemic} & $\bm{0.245}$  & $0.327$ & $0.782$ \\
\hline
\end{tabular}
\end{center}
\label{uav_brier_scores}
\end{table}

% \vspace*{-\baselineskip}

To compute the Brier score in this environment, the outcome we analyze is to have UAV fly to a target location in a given amount of time. We again analyze $250$ scenarios with different initial conditions. Brier score values are reported in Table \ref{uav_brier_scores}. The Brier scores for this environment are also the lowest for the residual VAE and highest for the probabilistic MLP. When comparing the low epistemic versus high epistemic scenarios, all three models have higher Brier score values for the high epistemic scenario. This is expected because we expect less accurate outcome prediction for a high uncertainty environment.

\section{CONCLUSIONS AND FUTURE WORK}\label{conclusions}

To create trustworthy competency-aware autonomous agents, we provide the autonomous agents with the capability to forecast their outcomes and uncertainties. A significant component of this is quantification of both epistemic and aleatoric uncertainties. While modeling of epistemic uncertainty can be achieved using state-of-the-art methods such as deep ensembling, aleatoric uncertainty modelling has to include effects of epistemic uncertainty. To achieve this, we showed how deep generative models in the form of CVAE can be trained exclusively on aleatoric uncertainty residual. This residual takes into account the samples from both the observed training data trajectory as well as from the ensemble model forecasted trajectories to actively incorporate the effects of epistemic uncertainty. We presented our results on the Pusher robot and UAV environment under varying levels of uncertainties. 

There are many possible avenues of extension for this work. First, we recognize that the model design process of CVAE may itself introduce some epistemic uncertainty, even though its impact may be significantly lower than the effects of systemic epistemic uncertainty that we address here. While communicating uncertainties to the user, this potential source of uncertainty also has to be addressed. Additionally, we plan to explore the ties between the uncertainty types, the forecasted outcome distributions, and the agent's performance. To do this, we will task the agent with multiple and increasingly complex objectives while studying the impacts of uncertainties on the forecasted and observed outcomes. As a longer term goal, we are looking to deploy our developed method on a real world UAV platform to demonstrate its utility beyond the simulated environment, thus leading to reliable and effective autonomous agents.

\addtolength{\textheight}{-3cm}

%%%%%%%%%%%%%%%%%%%%%%%%%%%%%%%%%%%%%%%%%%%%%%%%%%%%%%%%%%%%%%%%%%%%%%%%%%%%%%%%
% \section*{APPENDIX}

% Appendixes should appear before the acknowledgment.

\section*{ACKNOWLEDGMENT}

This material is based upon work supported by the Defense Advanced Research Projects Agency (DARPA) under Contract No. HR001120C0032. Any opinions, findings and conclusions or recommendations expressed in this material are those of the author(s) and do not necessarily reflect the views of DARPA.

%%%%%%%%%%%%%%%%%%%%%%%%%%%%%%%%%%%%%%%%%%%%%%%%%%%%%%%%%%%%%%%%%%%%%%%%%%%%%%%%

% References

\bibliographystyle{IEEEtran}
\bibliography{ref}

% Generated by IEEEtran.bst, version: 1.14 (2015/08/26)
\begin{thebibliography}{10}
\providecommand{\url}[1]{#1}
\csname url@samestyle\endcsname
\providecommand{\newblock}{\relax}
\providecommand{\bibinfo}[2]{#2}
\providecommand{\BIBentrySTDinterwordspacing}{\spaceskip=0pt\relax}
\providecommand{\BIBentryALTinterwordstretchfactor}{4}
\providecommand{\BIBentryALTinterwordspacing}{\spaceskip=\fontdimen2\font plus
\BIBentryALTinterwordstretchfactor\fontdimen3\font minus
  \fontdimen4\font\relax}
\providecommand{\BIBforeignlanguage}[2]{{%
\expandafter\ifx\csname l@#1\endcsname\relax
\typeout{** WARNING: IEEEtran.bst: No hyphenation pattern has been}%
\typeout{** loaded for the language `#1'. Using the pattern for}%
\typeout{** the default language instead.}%
\else
\language=\csname l@#1\endcsname
\fi
#2}}
\providecommand{\BIBdecl}{\relax}
\BIBdecl

\bibitem{trust_in_automation}
J.~D. Lee and K.~A. See, ``Trust in automation: Designing for appropriate
  reliance,'' \emph{Human Factors}, 2004.

\bibitem{buildingtrust}
S.~Ososky, D.~Schuster, E.~Phillips, and F.~Jentsch, ``Building appropriate
  trust in human-robot teams,'' \emph{AAAI Spring Symposium - Technical
  Report}, 2013.

\bibitem{Aitken2016AssurancesAM}
M.~Aitken, N.~R. Ahmed, D.~A. Lawrence, B.~Argrow, and E.~W. Frew, ``Assurances
  and machine self-confidence for enhanced trust in autonomous systems,''
  \emph{Robotics: Science and Systems (RSS) Workshop on Social Trust in
  Autonomous Systems}, 2016.

\bibitem{israelsen}
B.~W. Israelsen and N.~R. Ahmed, ``“{D}ave...{I} can assure you ...that
  it’s going to be all right ...” {A} definition, case for, and survey of
  algorithmic assurances in human-autonomy trust relationships,'' \emph{ACM
  Computing Surveys (CSUR)}, 2019.

\bibitem{acharya2022competency}
A.~Acharya, R.~Russell, and N.~R. Ahmed, ``Competency assessment for autonomous
  agents using deep generative models,'' \emph{IEEE/RSJ International
  Conference on Intelligent Robots and Systems (IROS)}, 2022.

\bibitem{ha2018worldmodels}
D.~Ha and J.~Schmidhuber, ``World models,'' in \emph{Conference on Neural
  Information Processing Systems (NIPS)}, 2018.

\bibitem{rosemaryke_longtermfuture}
N.~R. Ke, A.~Singh, A.~Touati, A.~Goyal, Y.~Bengio, D.~Parikh, and D.~Batra,
  ``Learning dynamics model in reinforcement learning by incorporating the long
  term future,'' \emph{International Conference on Learning Representations
  (ICLR)}, 2019.

\bibitem{chua2018deep}
K.~Chua, R.~Calandra, R.~McAllister, and S.~Levine, ``Deep reinforcement
  learning in a handful of trials using probabilistic dynamics models,''
  \emph{Conference on Neural Information Processing Systems (NIPS)}, 2018.

\bibitem{clements_rl_risks_uncertainty}
W.~R. Clements, B.~Robaglia, B.~V. Delft, R.~B. Slaoui, and S.~Toth,
  ``Estimating risk and uncertainty in deep reinforcement learning,''
  \emph{International Conference on Machine Learning (ICML) Workshop on
  Uncertainty and Robustness in Deep Learning}, 2020.

\bibitem{kendall_gal}
A.~Kendall and Y.~Gal, ``What uncertainties do we need in {B}ayesian deep
  learning for computer vision?'' in \emph{Conference on Neural Information
  Processing Systems (NIPS)}, 2017.

\bibitem{MacKay92bayesianmethods}
D.~J. MacKay, ``Bayesian methods for adaptive models,'' 1992.

\bibitem{MacKay95probablenetworks}
D.~J.~C. MacKay, ``Probable networks and plausible predictions -- a review of
  practical {B}ayesian methods for supervised neural networks,'' \emph{Network:
  computation in neural systems}, 1995.

\bibitem{neal_bayesian}
R.~M. Neal, \emph{Bayesian Learning for Neural Networks}.\hskip 1em plus 0.5em
  minus 0.4em\relax Berlin, Heidelberg: Springer-Verlag, 1996.

\bibitem{medicalceeplearning2}
A.~Esteva, A.~Robicquet, B.~Ramsundar, V.~Kuleshov, M.~DePristo, K.~Chou,
  C.~Cui, G.~Corrado, S.~Thrun, and J.~Dean, ``A guide to deep learning in
  healthcare,'' \emph{Nature medicine}, 2019.

\bibitem{medicaldeeplearning}
G.~J.~S. Litjens, T.~Kooi, B.~E. Bejnordi, A.~A.~A. Setio, F.~Ciompi,
  M.~Ghafoorian, J.~van~der Laak, B.~van Ginneken, and C.~I. S{\'a}nchez, ``A
  survey on deep learning in medical image analysis,'' \emph{Medical image
  analysis}, 2017.

\bibitem{roboticsdeeplearning}
H.~A. Pierson and M.~S. Gashler, ``Deep learning in robotics: a review of
  recent research,'' \emph{Advanced Robotics}, 2017.

\bibitem{roboticsdeeplearning2}
L.~Tai, J.~Zhang, M.~Liu, J.~Boedecker, and W.~Burgard, ``A survey of deep
  network solutions for learning control in robotics: From reinforcement to
  imitation,'' \emph{arXiv preprint arXiv:1612.07139}, 2016.

\bibitem{autdrivingdeeplearning}
S.~Grigorescu, B.~Trasnea, T.~Cocias, and G.~Macesanu, ``A survey of deep
  learning techniques for autonomous driving,'' \emph{Journal of Field
  Robotics}, 2020.

\bibitem{autdrivingdeeplearning2}
S.~Kuutti, R.~Bowden, Y.~Jin, P.~Barber, and S.~Fallah, ``A survey of deep
  learning applications to autonomous vehicle control,'' \emph{IEEE
  Transactions on Intelligent Transportation Systems}, 2020.

\bibitem{ensembles}
B.~Lakshminarayanan, A.~Pritzel, and C.~Blundell, ``Simple and scalable
  predictive uncertainty estimation using deep ensembles,'' \emph{Conference on
  Neural Information Processing Systems (NIPS)}, 2017.

\bibitem{evidentialnetwork}
B.~Charpentier, D.~Zügner, and S.~Günnemann, ``Posterior network: Uncertainty
  estimation without ood samples via density-based pseudo-counts,''
  \emph{Conference on Neural Information Processing Systems (NeurIPs)}, 2020.

\bibitem{evidentialnetwork2}
B.~Charpentier, O.~Borchert, D.~Z{\"{u}}gner, S.~Geisler, and
  S.~G{\"{u}}nnemann, ``Natural posterior network: Deep {B}ayesian predictive
  uncertainty for exponential family distributions,'' \emph{International
  Conference on Learning Representations (ICLR)}, 2021.

\bibitem{depeweg2017decomposition}
S.~Depeweg, J.~M. Hernández-Lobato, F.~Doshi-Velez, and S.~Udluft,
  ``Decomposition of uncertainty in {B}ayesian deep learning for efficient and
  risk-sensitive learning,'' 2017.

\bibitem{jain2021deup}
M.~Jain, S.~Lahlou, H.~Nekoei, V.~Butoi, P.~Bertin, J.~Rector-Brooks,
  M.~Korablyov, and Y.~Bengio, ``{DEUP}: Direct epistemic uncertainty
  prediction,'' \emph{arXiv preprint arXiv:2102.08501}, 2021.

\bibitem{osband2021epistemic}
I.~Osband, Z.~Wen, M.~Asghari, M.~Ibrahimi, X.~Lu, and B.~Van~Roy, ``Epistemic
  neural networks,'' \emph{arXiv preprint arXiv:2107.08924}, 2021.

\bibitem{eriksson2019epistemic}
H.~Eriksson and C.~Dimitrakakis, ``Epistemic risk-sensitive reinforcement
  learning,'' \emph{European Symposium on Artificial Neural Networks (ESANN)},
  2020.

\bibitem{sedlmeier2019uncertainty}
A.~Sedlmeier, T.~Gabor, T.~Phan, L.~Belzner, and C.~Linnhoff-Popien,
  ``Uncertainty-based out-of-distribution detection in deep reinforcement
  learning,'' \emph{International Symposium on Applied Artificial Intelligence
  (ISAAI)}, 2019.

\bibitem{wu2021uncertainty}
Y.~Wu, S.~Zhai, N.~Srivastava, J.~Susskind, J.~Zhang, R.~Salakhutdinov, and
  H.~Goh, ``Uncertainty weighted actor-critic for offline reinforcement
  learning,'' \emph{International Conference on Machine Learning (ICML)}, 2021.

\bibitem{explorationinRL}
\BIBentryALTinterwordspacing
T.~Yang, H.~Tang, C.~Bai, J.~Liu, J.~Hao, Z.~Meng, and P.~Liu, ``Exploration in
  deep reinforcement learning: A comprehensive survey,'' \emph{CoRR}, vol.
  abs/2109.06668, 2021. [Online]. Available:
  \url{https://arxiv.org/abs/2109.06668}
\BIBentrySTDinterwordspacing

\bibitem{osband2016risk}
I.~Osband, ``Risk versus uncertainty in deep learning: {B}ayes, bootstrap and
  the dangers of dropout,'' in \emph{Conference on Neural Information
  Processing Systems (NIPS) Workshop on Bayesian Deep Learning}, 2016.

\bibitem{rigter2021risk}
M.~Rigter, B.~Lacerda, and N.~Hawes, ``Risk-averse {B}ayes-adaptive
  reinforcement learning,'' \emph{Conference on Neural Information Processing
  Systems (NeurIPs)}, 2021.

\bibitem{neuneier1998risk}
R.~Neuneier and O.~Mihatsch, ``Risk sensitive reinforcement learning,''
  \emph{Conference on Neural Information Processing Systems (NIPS)}, vol.~11,
  1998.

\bibitem{ivanovic2022propagating}
B.~Ivanovic, Y.~Lin, S.~Shrivastava, P.~Chakravarty, and M.~Pavone,
  ``Propagating state uncertainty through trajectory forecasting,'' in
  \emph{IEEE International Conference on Robotics and Automation (ICRA)}, 2022.

\bibitem{acharya2022uncertainty}
A.~Acharya, R.~Russell, and N.~R. Ahmed, ``Uncertainty quantification for
  competency assessment of autonomous agents,'' \emph{IEEE International
  Conference on Robotics and Automation (ICRA) Workshop on Safe and Reliable
  Robot Autonomy under Uncertainty}, 2022.

\bibitem{huang2022evaluating}
Z.~Huang, H.~Lam, and H.~Zhang, ``Evaluating aleatoric uncertainty via
  conditional generative models,'' \emph{arXiv preprint arXiv:2206.04287},
  2022.

\bibitem{Kingma2014}
D.~P. Kingma and M.~Welling, ``{Auto-Encoding Variational {B}ayes},'' in
  \emph{International Conference on Learning Representations (ICLR)}, 2014.

\bibitem{condVAE}
K.~Sohn, H.~Lee, and X.~Yan, ``Learning structured output representation using
  deep conditional generative models,'' \emph{Conference on Neural Information
  Processing Systems (NIPS)}, 2015.

\bibitem{brockman2016openai}
G.~Brockman, V.~Cheung, L.~Pettersson, J.~Schneider, J.~Schulman, J.~Tang, and
  W.~Zaremba, ``Openai gym,'' \emph{arXiv preprint arXiv:1606.01540}, 2016.

\bibitem{conlon2022generalizing}
N.~Conlon, A.~Acharya, J.~McGinley, T.~Slack, C.~A. Hirst, M.~D'Alonzo, M.~R.
  Hebert, C.~Reale, E.~W. Frew, R.~Russell \emph{et~al.}, ``Generalizing
  competency self-assessment for autonomous vehicles using deep reinforcement
  learning,'' in \emph{American Institute of Aeronautics and Astronautics
  (AIAA) SciTech 2022 Forum}, 2022.

\end{thebibliography}

\end{document}